\documentclass[sigconf]{acmart}

\renewcommand\footnotetextcopyrightpermission[1]{} 
\AtBeginDocument{%
  \providecommand\BibTeX{{%
    \normalfont B\kern-0.5em{\scshape i\kern-0.25em b}\kern-0.8em\TeX}}}

\setcopyright{acmcopyright}
\copyrightyear{}
\acmYear{}
\acmDOI{}

\acmConference[]{}{}{}
\acmBooktitle{}
\acmPrice{}
\acmISBN{}

\begin{document}
\pagestyle{plain}
\title{Issues in Object Detection in Videos using Common Single-Image CNNs}


\author{Spencer Ploeger}
\affiliation{\institution{University of Guelph}}
\email{sploeger@uoguelph.ca}

\author{Lucas Dasovic}
\affiliation{\institution{University of Guelph}}
\email{ldasovic@uoguelph.ca}


\begin{abstract}
    A growing branch of computer vision is object detection. Object detection is used in many applications such as industrial process, medical imaging analysis, and autonomous vehicles. The ability to detect objects in videos is crucial. Object detection systems are trained on large image datasets. For applications such as autonomous vehicles, it is crucial that the object detection system can identify objects through multiple frames in video. There are many problems with applying these systems to video. Shadows or changes in brightness that can cause the system to incorrectly identify objects frame to frame and cause an unintended system response. There are many neural networks that have been used for object detection and if there was a way of connecting objects between frames then these problems could be eliminated.
    
 For these neural networks to get better at identifying objects in video, they need to be re-trained. A dataset must be created with images that represent consecutive video frames and have matching ground-truth layers. A method is proposed that can generate these datasets. The ground-truth layer contains only moving objects. To generate this layer, FlowNet2-Pytorch was used to create the flow mask using the novel Magnitude Method. As well, a segmentation mask will be generated using networks such as Mask R-CNN or Refinenet. These segmentation masks will contain all objects detected in a frame. By comparing this segmentation mask to the flow mask ground-truth layer, a loss function is generated. This loss function can be used to train a neural network to be better at making consistent predictions on video.
 
The system was tested on multiple video samples and a loss was generated for each frame, proving the Magnitude Method’s ability to be used to train object detection neural networks in future work.

\end{abstract}





\begin{teaserfigure}
\centering
  \includegraphics[width=0.8\textwidth]{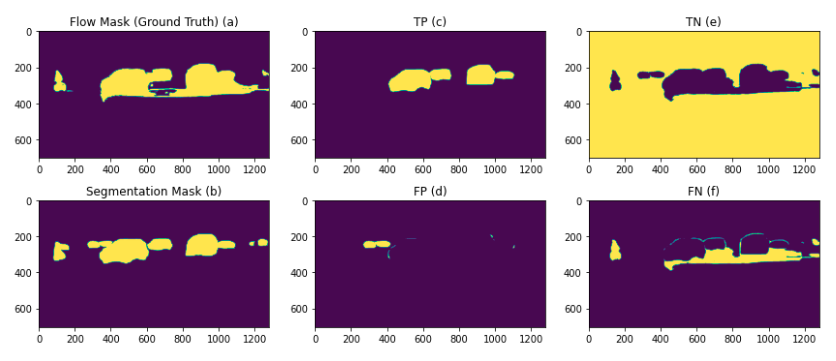}
  \caption{(a) The flow mask ground-truth layer, (b) A segmentation mask generated by Detectron2 using Mask R-CNN, (c) True Positive, (d) False Positive, (e) True Negative, (f) False Negative}
  \label{fig:teaser}
\end{teaserfigure}

\maketitle

\section{Introduction}
Object detection is part of a wider field called computer vision. Object detection is mainly concerned with identifying individual instances of an objects in an image. This is a very fast-growing branch of computer vision, and for good reason. There are many applications where object detection is crucial for an overall system to function. Important applications of object detection include industrial process control and monitoring, medical image analysis and target identification for military applications. One of the most important applications of object detection is for autonomous vehicles. Autonomous vehicles can use different object detection frameworks to identify hazards and obstacles in their surroundings and navigate around them accordingly. Most modern object detection frameworks are neural networks. Some modern and cutting edge networks used for object detection are Mask R-CNN \cite{mask_rcnn}, Faster R-CNN \cite{fasterrcnn} and Panoptic FPN \cite{panoptic}.

There is one thing in common with these networks, and that is that they are all trained on datasets of images, such as the COCO dataset \cite{coco} or Pascal VOC dataset \cite{voc}. These datasets are very large. As such, after training neural networks using these datasets, the networks are very good at identifying instances of objects in single frames. Identifying objects in single frames are useful for many applications, but not for the field of autonomous vehicle navigation. In these applications, the input to the neural networks would not be single frames, but instead high framerate video. The networks need to quickly run on each frame of the video, and return any objects detected as well as their positions.

While working very well on images, these previously mentioned networks do not produce smooth predictions on video. The root of this problem comes down to how the networks were trained. Since the networks were trained on single frames, they also make predictions on single frames. In other words, the input video split up and fed into the network frame by frame. The output is then stitched back together afterwards. 

\begin{figure}[H]
  \includegraphics[width=\linewidth]{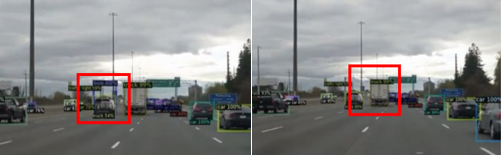}
  \caption{Left: A truck is identified as a van and a car at once. Right: The truck is not detected in the next frame}
  \label{fig:sporadic}
\end{figure}

The scenario becomes even more difficult for live video. In each frame of the live video input, there are often changes in brightness, contrast and shadowing. These frame-by-frame changes can cause the network to make odd, incorrect, or otherwise sporadic object predictions for a few frames, only to have these objects disappear as shadows or lighting change in the next frame. An example of a sporadic prediction is seen in Figure \ref{fig:sporadic}. These sporadic and incorrect predictions are called artifacts. This is clearly an issue for autonomous vehicles which navigate based on the predictions from networks like these. Imagine if an autonomous car was travelling at a high rate of speed, only to have an incorrectly detected obstacle appear in front of it out of error. What should the car do? How should it react? It is best to avoid this situation caused by artifact predictions.

If a neural network had some notion of connectivity of objects between frames, this issue of artifacting could be eliminated from the networks predictions. The network could be trained to not make an object prediction in a certain place unless the object was there in the previous frame. This would mean that a random shadow or change of brightness would not trigger an incorrect prediction, making the overall predictions in each frame smoother and more consistent.

\section{Related Work}

\subsection{Mask R-CNN}\label{maskrcnn}

\begin{figure}[H]
  \includegraphics[width=0.8\linewidth]{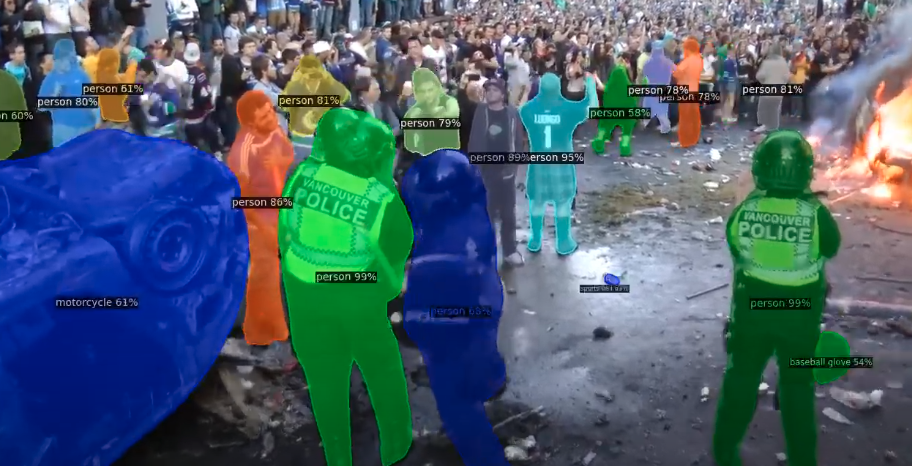}
  \caption{Example of predictions made by Mask R-CNN} 
  \label{fig:maskrcnn_example}
\end{figure}

One state-of-the-art object detection framework is called Mask R-CNN \cite{mask_rcnn}. It extends a history of region-based neural network detectors originally proposed by Ross Girshick in 2015. \cite{fast} It combines both object detection and semantic segmentation into one prediction for a whole image. \cite{mask_rcnn} This means it is particularly good at instance segmentation, which is a crucial task for autonomous vehicle navigation. It works as essentially what is a two-step process.\cite{deeplearning} \cite{mask_rcnn} The first step is that a fully-convolutional network (FCN) generates regions of interest for where objects are likely to be in an image. Then a region proposal network (RPN) generates masks and labels for each object in the image. These two separate networks within the larger framework share some convolutional layers, but for the most part they function independently \cite{mask_rcnn}. This process leads to very, very good predictions on single images. The output consists of an object masks, labels and a score of certainty for each prediction. A visualized example of a Mask R-CNN prediction is seen in Figure \ref{fig:maskrcnn_example}.

\subsection{Refinenet}

Another very recently developed network is called Refinenet. Refinenet differs in a few ways from Mask R-CNN. Firstly, Refinenet is a fully convolutional network. \cite{refinet} It does not have a two-step approach like Mask R-CNN. \cite{deeplearning} This means that the output from the network is already a mask. There is no second stage of processing which needs to be done to it (like the RPN does in Mask R-CNN). Essentially, the network generates a few scaled-down versions of the input image. It uses a Residual Conv Unit (RCU) to generate feature maps on each scaled-down image, combines these feature maps into pooling blocks, and then another RCU generates output.\cite{refinet} A modified version of this network is called Lightweight Refinenet. \cite{lightweight} This modified network is more compact that the original Refinenet network. Lightweight Refinenet does away with the RCU, and replaces all 3x3 convolutions with 1x1 convolutions. This makes lightweight Refinenet much faster \cite{fast} and therefore more suited to processing live video, such as on autonomous vehicles. Lightweight Refinenet generates masks, similar to those generated from Mask R-CNN. These masks identify the class which the object belongs to.

\subsection{Detectron2}
Detectron2 is a computer vision framework designed by Facebooks AI Research group. \cite{detectron} It is capable of implementing various computer vision frameworks, such as Mask R-CNN as mentioned above. It also provides helper functions to make tasks like visualization and training easier. In this context, Detectron2 is used to apply Mask R-CNN to sample images and videos.

\subsection{FlowNet 2.0}

The idea of optical flow has been used since the 1980’s and has been used to improve the accuracy and power consumption of various computer vision models when classifying videos. \cite{optical} Optical flow is a per pixel prediction that assumes a brightness constancy, and pixels characteristics are predicted by a flow field at different locations. \cite{optical} One implementation of optical flow is FlowNet 2.0, which uses a convolutional CNN architecture to directly learn the concept of optical flow. \cite{flownet2} FlowNet 2.0 is based off of four different algorithms, FlowNetS, FlowNetC, FlowNetCSS, and FlowNetSD. The FlowNet 2.0 architecture uses these different algorithms to detect large or small displacements between two images using brightness error to calculate movement between images. \cite{optical} The Pytorch code base is used to generate a flow field that can be used to see how movement of pixels is changing between images. FlowNet2 data can be visualized to determine the movement of an object. Visualization works by transforming each pixels flow vector (U,V) to (X,Y) and putting it on a Cartesian plane. The scale of the vectors are determined each frame relative to all other pixels with the only restriction being the magnitude of vectors must be less than $1x10^9$. This Cartesian vector is then translated to a colour based off the scale in Figure~\ref{fig:flow_scale} below.

\begin{figure}[H]
  \includegraphics[width=0.8\linewidth]{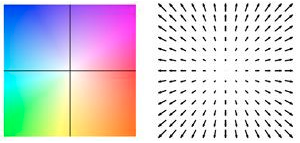}
  \caption{Flow scale with colour map and vector demonstration \cite{flowscale}} 
  \label{fig:flow_scale}
\end{figure}

\begin{figure}[H]
  \includegraphics[width=0.8\linewidth]{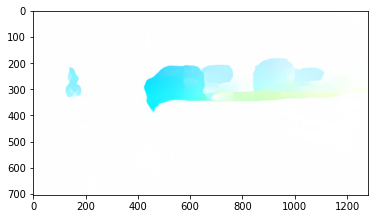}
  \caption{Example of FlowNet output on a frame of a video} 
  \label{fig:flow_example}
\end{figure}

\section{Approach}

To allow object detection neural networks such as Refinenet and Mask R-CNN to get better at identifying objects in video without introducing artifacts, as previously mentioned, there needs to be a way to re-train these networks. The networks must be re-trained using a set of data that is known to not contain any of these artifacts. Additionally, this theoretical dataset should accurately represent consecutive video frames, as this is the desired use case and main motivation for re-training these networks. In other words, there must be a way to create a dataset of images that represent consecutive video frames, and that have matching ground-truth layers which are free of artifacts. Once this dataset is obtained, the existing neural networks can be re-trained, and evaluated for better performance on video. By re-training using a dataset that does not contain any artifacts, anytime the detection network does make an incorrect, sporadic detection, this will ideally be represented in the loss function, and result in the network being adjusted accordingly.

\subsection{Source of Training Data}

As previously mentioned, the training dataset should represent consecutive video frames as much as possible. Therefore, it is proposed that the new training dataset be created directly from sample videos. This is both a practical, and fast approach. To separate a desired sample video into individual frames, the command line tool FFmpeg \cite{ffmpeg} is used. To create a dataset, the first step is to feed the sample video into FFmepg. The output from FFmpeg is a set of frames that make up the original video. These frames are directly representative of the consecutive, single frames which these networks must run predictions on when they are fed video.

\subsection{Generating the Ground-Truth Layer with FlowNet 2.0}

The next step is to take the previously generated frames that will make up the dataset and generate a ground-truth layer from them. This layer must contain all objects in this frame which are moving. Random artifacts incorrectly predicted by the object detection networks are not moving, as they just appear in one frame and disappear in the next. By having the ground-truth layer only include moving objects, if the object detection network did incorrectly detect an artifact, this would be represented in the loss function. From this, the network will be modified as to not predict these single, stationary artifacts.

To create the ground-truth layer, FlowNet2-Pytorch will be used. \cite{flowpytorch} This layer will be called the flow mask. This flow mask will be a binary mask that represents all movement in a given frame. Like previously explained, FlowNet2-Pytorch is an implementation of the FlowNet2 optical flow estimation framework that implements PyTorch. Using FlowNet2 is a novel approach since it allows for the automatic generation of ground-truth masks on the dataset frames. It is also convenient that it runs on a common GPU, much like the neural networks we are trying to improve. Essentially, FlowNet2 takes in a set of images and predicts the motion of each pixel in each frame. This motion prediction is in the form of a U and V vector. This vector gives the X and Y motion of the pixel on a Cartesian plane. After splitting up the frames of a sample video using FFmepg, the frames can be run through FlowNet2 to generate motion predictions for each frame. Flow data can be visualized with external libraries such as \textit{flow2image}.\cite{skeeet_2019} An example of this visualized flow data for a frame can be seen in Figure \ref{fig:flow_example}.

\subsection{Creating the Mask: The Magnitude Method}

The ground-truth layer must be created from the flow data. By doing so, there is a way to determine when artifact predictions have been made, and have them reflected by the loss function for that frame. A method is proposed for generating the ground truth layer using the flow data obtained from running FlowNet2 on each frame. This method is called the Magnitude Method. As previously explained, FlowNet2 generates a motion vector for each pixel. The magnitude of motion for each pixel can be calculated by:

\begin{equation}
    Mag = \sqrt{(U^2 + V^2)}
\end{equation}

This is an efficient and quick operation to perform for a whole frame of flow information that can be performed on the GPU. This way, the magnitude of motion for each pixel in a frame can be quickly determined. A threshold is arbitrarily set, using a guess and check method to find a good value. Using this magnitude information, a binary mask is created. Any pixels whose motion magnitude is greater than the threshold is present in the mask. An example of one of these masks generated by the magnitude method can be seen in Figure \ref{fig:teaser}(a). This mask is called the flow mask, and it is very significant. It gives a way of identifying moving objects that are present over time in consecutive frames. In other words, there is now a way to relate objects in one frame to the next. With this valuable mask, it is possible to identify incorrectly detected artifacts. This mask will be used as the ground-truth layer when re-training existing networks.

\subsection{Comparing with Existing Models}

To compare with existing object detection networks like Mask R-CNN or Refinenet, there must be a way to generate masks from the output of these networks to compare against. Obviously object detection networks like Refinenet or Mask R-CNN can produce and masks around objects, as that is their main purpose. But this is not what we are concerned about for this purpose. Here, exactly what types of objects which are detected by these networks is not important. All that matters is the total mask of detected objects. This is because we are not trying to improve specific object detection, but rather just trying to decrease incorrect and sporadic detections. We need a segmentation mask that represents all instances detected for a single frame, all in a single mask. This is what can be compared to the ground-truth layer described above. We generate this binary segmentation mask by performing a logical OR on the mask for each object detected in a frame. This gives one overall frame that has all detected objects for a frame in one mask. An example of this type of segmentation mask can be seen in Figure \ref{fig:teaser}(b). 

Any objects that are present in the segmentation mask that are not present in the ground-truth flow mask are likely incorrectly predicted artifacts. By comparing the flow masks for each frame of a video with the matching ground-truth layer (flow mask), we can calculate the true positives, true negatives, false positives, and false negatives for each frame. This is are shown in Figure \ref{fig:teaser}(c)-(f). The intersection-over-union (IOU) for each frame can be calculated by:

\begin{equation}
    IOU = \frac{TP}{(TP + FP + FN)}
\end{equation}

Similarly, the loss can be calculated as:

\begin{equation}
    Loss = 1 - IOU
\end{equation}

Using the loss equation above, a loss can be generated for each frame in a given video clip. This loss function can be used to re-train the original object detection network, such as Mask R-CNN or Refinenet. As part of this study, some research was done into actually re-training the original object detection networks. Re-training a network like Refinenet would prove to be much easier than re-training Mask R-CNN. Refinenet is a purely convolutional network. This means that the output mask comes directly from the network layers. \cite{refinet} In other words, there is no post-processing of the masks before they are output. If we desired to go modify the network, it would be very easy as the layers of the network are directly exposed. On the other hand, Mask R-CNN is a type of multi-step network. \cite{deeplearning} The output from Mask R-CNN is not the direct result of convolution operations. Instead, there are multiple steps involved in getting an output. These steps are described in section. \ref{maskrcnn} Because of these extra steps, it is more difficult to work backwards in the networks layers to re-train it. 

\section{Results}

This study did not go as far as to re-train the original object detection networks with the obtained loss function. Rather than using the generated loss functions to retrain the object detection networks, the above process was applied to two separate networks. The goal was to investigate the feasibility in generating a loss function as previou sly described. Table~\ref{tab:average} below shows the average loss for 30 frames of three sample videos. The loss was calculated for both Refinenet  as well as Mask R-CNN. It is clear that the proposed method is capable of calculating loss between the ground-truth flow mask and the predicted segmentation mask. These results are demonstrated in the table below.

\begin{table}[H]
\caption{Average loss for 30 frames of three sample videos}
\label{tab:average}
\begin{tabular}{|l|c|c|}
\hline
\multicolumn{1}{|c|}{Sample Video:} & \begin{tabular}[c]{@{}c@{}}Average Loss - \\ Refinenet\end{tabular} & \begin{tabular}[c]{@{}c@{}}Average Loss - \\ Mask R-CNN\end{tabular} \\ \hline
driveby.mp4                         & 0.7651                                                              & 0.7306                                                               \\ \hline
walking.mp4                         & 0.8396                                                              & 0.8460                                                               \\ \hline
driving\_on\_401.mp4                & 0.8904                                                              & 0.9075                                                               \\ \hline
\textbf{Average}                    & \textbf{0.8317}                                                     & \textbf{0.8280}                                                      \\ \hline
\end{tabular}
\end{table}

\section{Discussion}
The average loss for both models tested is almost the same (within 0.01 of each other). This indicates that the Magnitude Method for creating flow masks, and using these as ground-truth layers is equally valid for different types of neural networks.

\begin{figure}[H]
\centering0
  \includegraphics[width=0.9\linewidth]{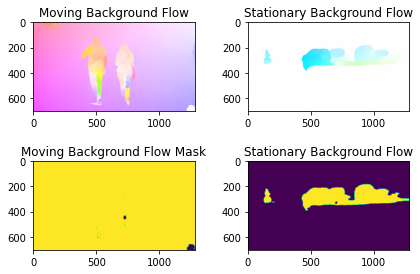}
  \caption{Left upper and lower: A moving background makes it harder to differentiate the flow mask from the background using the magnitude method, Right upper and lower: A stationary background makes it much easier to differentiate the background from foreground objects when creating the flow mask.}
  \label{fig:moving_background}
\end{figure}

We notice the loss is much higher on the walking and driving videos (walking.mp4 and driving\_on\_401.mp4). This is interesting to note because they have a major difference compared to the driveby video, driveby.mp4. The driveby video has a stationary background, with many objects moving in the foreground. When visualizing its flow data, there is a clear distinction between the stationary background and moving foreground. This means that when the Magnitude Method is applied to it, the resulting flow mask also has a clear distinction between the moving foreground and stationary background. Conversely, the walking and driving videos both contain a moving background (relative to the camera) and a moving foreground (people and cars, respectively). When visualizing the flow data for it, there is not as clear of a distinction between foreground and background. This becomes evident in Figure~\ref{fig:moving_background}. From this figure we can see that for the image pair on the left, both the background and foreground are moving. As such, they both have a magnitude and direction of movement, that is why they are coloured. When the flow mask is generated using the Magnitude Method, the whole image's motion magnitude is greater than the threshold, and that is why the flow mask becomes the whole frame (bottom left). For the right side image pair of Figure \ref{fig:moving_background}, the background of the flow is white, which means it is not moving according to the scale in Figure \ref{fig:flow_scale}. The background has a much lower magnitude of movement than the cars, which appears blue. The magnitude of the car movement is greater than the threshold, so they appear in the flow mask while the background does not(bottom right).

The consequences of this are that the Magnitude Method for creating flow masks to use as the ground-truth layer is only valid for videos where the background is stationary relative to the camera.

For these instances where the Magnitude Method produces acceptable flow masks to use as the ground-truth layer, the method seems valid. In these cases where an adequate and clear flow mask can be created, it gives a good representation of what is moving in a frame and what is not. This can successfully be compared with the segmentation mask for the same frame, and a loss can be calculated. This loss function should be able to be used to train the original predictor network to better and more smoothly make predictions on videos.

\section{CONCLUSION}
There is a clear need for reliable object detection neural networks that can reliably run on live video for applications such as autonomous vehicles. As these networks were trained on single frame images, they do not perform consistently well on videos. As such, a method was proposed to improve this. The method involved creating a flow mask by applying the Magnitude Method to flow data for a set of images. A segmentation mask was also generated that contained every detected object for each frame. Then from these two masks, an IOU and loss was calculated for each frame. This loss can be used to directly re-train simple purely convolutional networks like Refinenet. This method is valid for any video in which the background is not stationary relative to the camera.


\bibliographystyle{ACM-Reference-Format}
\bibliography{sample-base}



\end{document}